\def\year{2022}\relax
\documentclass[letterpaper]{article} 
\usepackage{aaai22}  
\usepackage{times}  
\usepackage{helvet}  
\usepackage{courier}  
\usepackage[hyphens]{url}  
\usepackage{graphicx} 
\usepackage{natbib}  
\usepackage{caption} 
\DeclareCaptionStyle{ruled}{labelfont=normalfont,labelsep=colon,strut=off} 
\usepackage{algorithm}
\usepackage{algorithmic}
\usepackage{xspace}
\usepackage{tabularx}
\usepackage{multirow}
\usepackage{amsmath}
\usepackage{amsfonts}
\usepackage{bm}
\usepackage{bbding}
\usepackage[symbol]{footmisc}

\newcommand{\bleu}{\textsc{Bleu}\xspace}
\newcommand{\rone}{\textsc{R@1}\xspace}

\newcommand{\meteor}{\textsc{Meteor}\xspace}
\newcommand{\rouge}{\textsc{Rouge}\xspace}

\newcommand{\roberta}{\textsc{RoBERTa}\xspace}
\newcommand{\robertawd}{\textsc{RoBERTa-WD}\xspace}
\newcommand{\robertawdd}{\textsc{RoBERTa-WD2}\xspace}

\newcommand{\robertawdmtl}{\textsc{RoBERTa-WD-MTL}\xspace}

\usepackage{newfloat}
\usepackage{listings}
\floatstyle{ruled}
\newfloat{listing}{tb}{lst}{}
\floatname{listing}{Listing}
\title{Towards Generalized Models for Task-oriented Dialogue Modeling on Spoken Conversations}
\author {
    Ruijie Yan\textsuperscript{\rm 1}\footnote{Work done during an internship at Ant Group.}, 
    Shuang Peng\textsuperscript{\rm 2}, 
    Haitao Mi\textsuperscript{\rm 2}, 
    Liang Jiang\textsuperscript{\rm 2}, 
    Shihui Yang\textsuperscript{\rm 2}, 
    Yuchi Zhang\textsuperscript{\rm 2}, 
    Jiajun Li\textsuperscript{\rm 2}, 
    Liangrui Peng\textsuperscript{\rm 2}, 
    Yongliang Wang\textsuperscript{\rm 2}, 
    Zujie Wen\textsuperscript{\rm 2}
}
\affiliations {
    \textsuperscript{\rm 1} Beijing National Research Center for Information Science and Technology\\
    \textsuperscript{\rm 1} Department of Electronic Engineering, Tsinghua University, Beijing, China \\
    \textsuperscript{\rm 2} Ant Group\\
    yrj17@mails.tsinghua.edu.cn, penglr@tsinghua.edu.cn \\ 
    \{jianfeng.ps, haitao.mi, tianxuan.jl, yilin.ysh, yuchi.zyc, suojue.ljj, yongliang.wyl, zujie.wzj\}@antgroup.com
}

\usepackage{bibentry}

\begin{document}

\maketitle

\begin{abstract}
Building robust and general dialogue models for spoken conversations is challenging due to the gap in distributions of spoken and written data. This paper presents our approach to build generalized models for the Knowledge-grounded Task-oriented Dialogue Modeling on Spoken Conversations Challenge of DSTC-10. In order to mitigate the discrepancies between spoken and written text, we mainly employ extensive data augmentation strategies on written data, including artificial error injection and round-trip text-speech transformation. To train robust models for spoken conversations, we improve pre-trained language models, and apply ensemble algorithms for each sub-task. Typically, for the detection task, we fine-tune \roberta and ELECTRA, and run an error-fixing ensemble algorithm. For the selection task, we adopt a two-stage framework that consists of entity tracking and knowledge ranking, and propose a multi-task learning method to learn multi-level semantic information by domain classification and entity selection. For the generation task, we adopt a cross-validation data process to improve pre-trained generative language models, followed by a consensus decoding algorithm, which can add arbitrary features like relative \rouge metric, and tune associated feature weights toward \bleu directly. Our approach ranks third on the objective evaluation and second on the final official human evaluation.
\end{abstract}

\setlength{\belowcaptionskip}{-4pt}

\section{Introduction}

\begin{figure*}
  \centering
  \includegraphics[width=.975\textwidth]{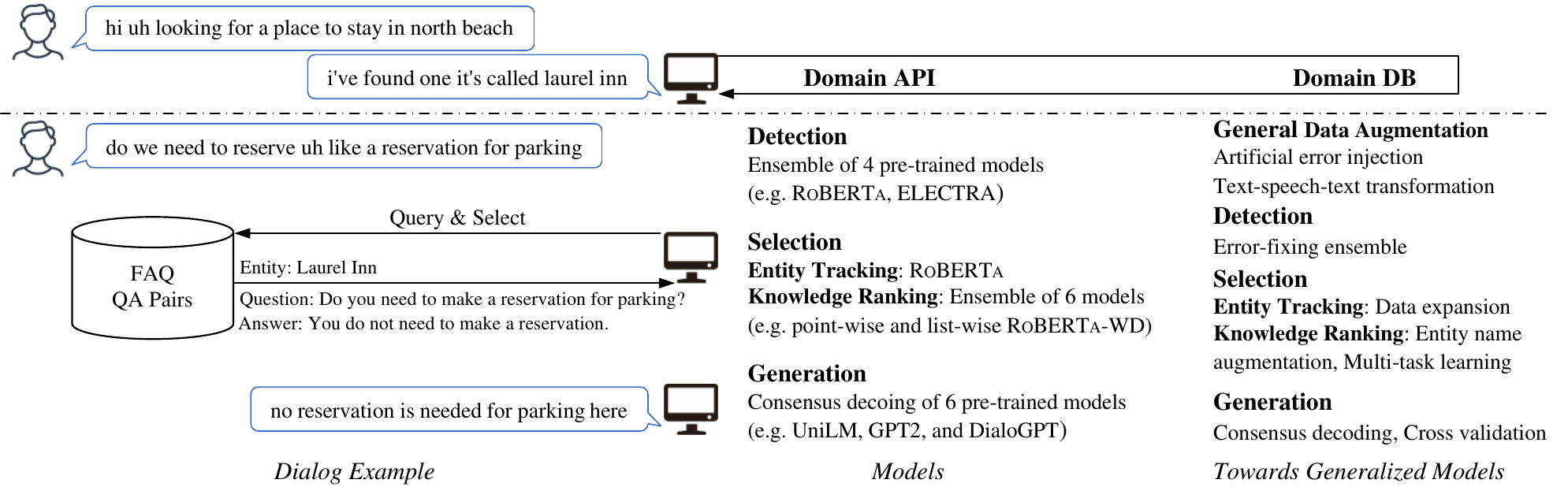}
  \abovecaptionskip=0pt
  \caption{An example and overview of our approach on detection, selection and generation tasks. 
    The selection task consists of two sub-tasks: entity tracking and knowledge ranking.
    Entity tracking aims to find a list of entities mentioned in a dialogue, while
    knowledge ranking ranks a list of knowledge, 
    and returns the top-$n$ related knowledge.}
  \label{fig:dialog_ex}
\end{figure*}

Although promising results have been achieved by dialogue systems on written conversations, using them directly on spoken conversations is challenging due to the differences in data distribution, including the discrepancy between writing and speaking, and the extra noises from speech recognition errors. In Dialog System Technology Challenges 10 (DSTC-10), the sub-track 2 of the ``Knowledge-grounded Task-oriented Dialogue Modeling on Spoken Conversations'' task proposes such a challenge, in which models are evaluated on spoken conversations while no spoken training data is provided. Therefore, it is crucial to build robust dialogue models with high generalization ability. 

The task extends the DSTC-9 track 1~\cite{kim2020domain} from written conversations to spoken conversations, where cross-domain dialogue agents are built to 
answer questions that cannot be solved with only domain APIs. Instead, system agents have to retrieve related question-answer (QA) pairs from an unstructured FAQ database, and generate a natural response based on the retrieved QA pair(s).
Thus, our tasks include (1) finding knowledge-seeking turns; (2) returning ranked QA pairs for each knowledge-seeking turn; and (3) generating a system response given QA pairs from task (2) and the dialogue history.

In this paper, we build generalized models for the task in the following ways.
First, to bridge the gap between writing and speaking, we employ various data augmentation methods to expand the training set, including artificial error injection and round-trip text-speech transformation. 
Second, we use pre-trained language models (e.g. \roberta~\cite{roberta}, ELECTRA~\cite{clark2020electra}, and UniLM~\cite{unilm}), and design different ensemble algorithms for each sub-task.
Third, for the selection task, we propose a multi-task learning mechanism to enhance models' ability to learn multi-level semantic information. 
We also introduce artificial sparse features to explicitly capture informative attributes of the candidate knowledge.
For the generation task, we mainly follow Mi et al.~\shortcite{mi2021towards}, and directly use their online sampling, and consensus decoding algorithms~\cite{pauls2009consensus}. 

In particular, we extend the work of Mi et al.~\shortcite{mi2021towards}, and make the following extra contributions in this paper. 
\begin{itemize}
    \item \textbf{Data augmentation.} We augment written data by injecting artificially-generated errors based on phonetic similarity, converting the original texts into sound waves by a text-to-speech (TTS) model and then transforming back into texts by an automated speech recognition (ASR) model, and splitting or inserting entity names in dialogues.
    \item \textbf{Multi-task learning.} We propose a multi-task learning method for knowledge ranking, in which a domain classification task and an entity selection task are assigned to learn multi-level semantic information.
    \item \textbf{Incremental improvements.}  We apply more pre-trained language models, e.g. ELECTRA, to increase the modeling diversity. 
          We also introduce additional artificial sparse features to explicitly capture informative attributes of knowledge snippets.
\end{itemize}

Our system achieves the third best in the official objective evaluation, and the second best on human evaluation.

\section{Task Description and Our Approach}
Task-oriented dialogue systems provide information to or complete actions for users through a natural language dialogue flow.
To build smart assistants that are not constrained to a set of pre-defined operations (APIs) and structural data (DBs), in the ``Beyond Domain APIs'' task of DSTC-9~\cite{kim2021beyond}, dialogue systems with high performance are proposed to utilize unstructured FAQ pairs to complete knowledge seeking tasks and generate natural responses~\cite{mi2021towards,he2021learning,tang2021radge}.

While in DSTC-9 we only focused on written conversations, the ``Knowledge-grounded Task-oriented Dialogue Modeling'' challenge in DSTC-10 extends the task to spoken conversations. No additional spoken training data is provided. As many machine learning methods work well only under the assumption that the training and test data are drawn from the same feature space and the same distribution, the generalization ability of models trained on the DSTC-9 dataset is unsatisfactory on spoken data. 

We adopt the method proposed by Mi et al.~\shortcite{mi2021towards} as our baseline system. 
Figure~\ref{fig:dialog_ex} gives a brief overview of the approach. The system first determines whether the utterance is a knowledge-seeking turn (\textbf{Detection}), then tracks entities in the dialogue history and ranks relevant QA pairs (\textbf{Selection}), and finally generates a natural response based on the retrieved QA pair (\textbf{Generation}).

We divide the selection task into two sub-tasks: \textbf{Entity Tracking} and \textbf{Knowledge Ranking}. Entity tracking aims to find a list of entities that are mentioned in dialogue. With these entities, we collect all knowledge associated with them, and knowledge ranking ranks those knowledge documents and returns top-$n$ knowledge. 

In order to build generalized models, 
we augment training data to simulate spoken dialogues. The current state-of-the-art pre-trained language models are employed for each task, followed by ensemble algorithms. 






\section{Key Components of Our Approach}
\subsection{Task Formulation}
For simplicity, a dialogue $D$ consists of a sequence of user and system utterances.
Let ${U_i}$ and ${S_i}$ represent a user utterance and a system utterance at turn $i$ respectively.

A knowledge set $\mathbf{K}$ is a list of knowledge, 
and each knowledge $k_j$ is a tuple of $\langle e_j, q_j, a_j \rangle $, 
where $e_j$ is an entity name or a domain name (for domain-level knowledge), 
$q_j$ is a knowledge question, and $a_j$ is a knowledge answer. 
We use $\mathbf{E}$ to denote a complete set of all $e_j$ in $\mathbf{K}$.
At each knowledge-seeking turn $i$, a ranked knowledge list $K^R_i$ is associated with this turn.
Let $k^n_i$ represent the $n$-th knowledge.

Given above annotations, we define our tasks as
\begin{itemize}
  \item \textbf{Detection} detects whether turn $i$ is a knowledge seeking turn given the context of $U_1$  ... $S_{i-1} U_{i}$.  
  \item \textbf{Selection} returns a top-$n$ ranked knowledge list $K^R_i$ at knowledge seeking turn $i$ from the whole knowledge set $\mathbf{K}$, 
        we divide this task into following two sub-tasks:
    \begin{itemize}
       \item \textbf{Entity Tracking} extracts a list of entities $E_i$, in which each entity $e_i$ is mentioned in the current dialogue;
       \item \textbf{Knowledge Ranking} first collects a candidate knowledge list $K^C_i$ based on entities in $E_i$, 
                         then ranks $K^C_i$ and returns top-$n$ knowledge $K^R_i$ ($k^1_i$, $k^2_i$, ... $k^n_i$). 
    \end{itemize}
  \item \textbf{Generation} predicts a system response $S^*_i$ given $K^R_i$ and the context of $U_1$  ... $S_{i-1} U_{i}$.
\end{itemize}

Besides annotations defined above, we also add special tags to mark the type of each block in data representation 
and separate the input in our models:
\begin{itemize}
    \item $\langle user \rangle$: start of a user utterance;
    \item $\langle sys \rangle$: start of a system utterance;
    \item $\langle kng \rangle$: start of a knowledge;
    \item $\langle kng_k \rangle$: start of the $k$-th best knowledge;
    \item $\langle ent \rangle$: start of an entity or a domain name;
    \item $\langle ans \rangle$: start of a knowledge answer;
    \item $\langle resp \rangle$: start of a response.
\end{itemize}

\begin{table*}
\begin{center}
\scalebox{0.90}{
\begin{tabular}{c|c|c|c|c}
\hline 
Sub-task & \multicolumn{2}{c|}{Context} & Target & Type   \\
\hline
Entity &  Sentence 1  & Sentence 2  & True/False & \multirow{2}{*}{Binary} \\\cline{2-4}
Tracking & $\langle user \rangle U_1$  ... $\langle sys \rangle S_{i-1} \langle user \rangle U_{i}$ & $\langle ent \rangle e_j$ & True &  \\
\hline
 & Sentence 1  & Sentence 2 & True/False & \multirow{2}{*}{Binary} \\\cline{2-4}
Knowledge & $\langle user \rangle U_1$  ... $\langle sys \rangle S_{i-1} \langle user \rangle U_{i}$ & $\langle kng \rangle q_{j} \langle ans \rangle a_{j}$ & True &  \\\cline{2-5}
Ranking     & Sentence 1  & Sentence 2 & One hot & \multirow{2}{*}{Multi-class} \\\cline{2-4}
 & $\langle user \rangle U_1$  ... $\langle sys \rangle S_{i-1} \langle user \rangle U_{i}$ & $\langle kng \rangle q^1_{j} \langle ans \rangle a^1_{j}$, ... , $\langle kng \rangle q^5_{j} \langle ans \rangle a^5_{j}$ & [0, 1, 0, 0, 0] &  \\
\hline
\end{tabular}
}
\end{center}
\abovecaptionskip=3pt
\caption{The data representations for entity tracking and knowledge ranking tasks. 
In the knowledge ranking task, we first use a point-wise \roberta to select top-$5$ related knowledge (rows 4 and 5), 
then, we develop a list-wise \roberta (the last two rows) to rank top-$5$ knowledge again, 
$\langle kng \rangle q^1_{j} \langle ans \rangle a^1_{j}$, ... , $\langle kng \rangle q^5_{j} \langle ans \rangle a^5_{j}$ means a batch of top-$5$ knowledge, and 
the objective function is to minimize the cross-entropy loss between the true distribution and the system prediction of five classes.
}
\label{tab:sel_task_dr}
\end{table*}

\subsection{Data Augmentation}\label{sec:da}

To alleviate the gap between writing and speaking, for each dialogue in the DSTC-9 dataset, we use the following three strategies to augment the utterances.
\begin{itemize}
    \item \textbf{Artificial error injection}. We simulate speech recognition errors by randomly replacing words based on phonetic similarity~\cite{ailsa2019effect}. The phonetically-similar alternatives are selected by approximate nearest neighbors search with angular distance. 
    The proportion of replaced words is sampled from 0.1 to 0.3.
    \item \textbf{Text-speech-text transformation}. We synthesize sound waves from the original texts using the Tacotron 2 TTS model~\cite{shen2018natural}, and then adopt the Deep Speech 2 ASR model~\cite{amodei2016deep} to recover texts from sound waves.
    \item \textbf{Entity name augmentation}. Entity name is one of the most important information for the selection task. People tend to omit certain words when they speak, and errors may occur when one or more words are omitted from the entity name. In addition, non-ground-truth entities that appeared in the conversation are easily identified as false positives.
    To alleviate this issue, we augment entity names in conversations according to the attributes of the candidate knowledge. 
    On one hand, if the candidate knowledge is positive and the corresponding entity name can be matched in the dialogue, 
    we split the entity name into two parts, and randomly move one of them to another place in the dialogue. 
    We also delete each word in the entity name with 
    a probability of 0.1. 
    On the other hand, if the candidate knowledge is negative and the corresponding entity name does not appear in the conversation, we randomly insert the entity name into the dialogue.
\end{itemize}

Table~\ref{tab:data_aug} shows some examples for data augmentation. For entity name augmentation, we use ``SW Hotel'' as an example of the negative entity name. 
Although the augmented entity name seems unnatural in the utterance, it requires the model to understand the semantic information in the dialogue rather than simply using keywords to match knowledge. 
In general, errors from artificial error injection are often moderate, while text-speech-text transformation simulates more challenging situations.
We apply artificial error injection and text-speech-text transformation for all three tasks 
on the whole DSTC-9 dataset to expand training data, 
while we only run entity name augmentation for the selection task 
with a probability of 0.3 along with the training process.

\begin{table}[ht]
    \centering
    \scalebox{0.90}{
    \begin{tabular}{ll}
        \hline
        Augmentation                    & Text \\
        \hline
        None & can I cooking at Hamilton lodge \\
        Error injection & can I \textbf{booking} at Hamilton \textbf{launch} \\
        Text-speech-text & can I cooking \textbf{and high museum large} \\
        Entity name (pos) & can I cooking \textbf{lodge} at Hamilton    \\
        Entity name (neg) & can I \textbf{SW Hotel} cooking at Hamilton lodge \\
        \hline
    \end{tabular}
    }
    \caption{Examples of different data augmentation methods. The augmented words are marked in \textbf{bold}. For entity name augmentation, ``pos'' and ``neg'' denote the candidate knowledge is positive or negative, and we take ``SW Hotel'' as an example of the negative entity name.}
    \label{tab:data_aug}
\end{table}

\subsection{Detection}\label{sec:det}

\subsubsection{Models and Data Representations}
Following Mi et al.~\shortcite{mi2021towards}, we treat detection as a \textit{binary} classification problem
and make use of pre-trained language models, such as \roberta~\cite{roberta} and ELECTRA~\cite{clark2020electra}. 
We represent the source side as a concatenation of all dialogue utterances with tag labels, i.e. the dialogue history:
\begin{itemize}
    \item[] $\langle user \rangle U_1$  ... $\langle sys \rangle S_{i-1} \langle user \rangle U_i$.
\end{itemize}
The target side is `True' or `False' for binary classification.

Our ensemble algorithm directly uses the error-fixing ensemble of Mi et al.~\shortcite{mi2021towards}.


\subsection{Selection}\label{sec:sel}
Following Mi et al.~\shortcite{mi2021towards}, we split selection into two sub-tasks: entity tracking and knowledge ranking, 
and treat these tasks as \textit{sentence pair} classification problems. Data representation for selection includes three parts in Table~\ref{tab:sel_task_dr}.
\begin{itemize}
    \item Sentence 1: the history of a dialogue,
    \item Sentence 2: an entity name or a knowledge for entity tracking or knowledge ranking respectively,
    \item Target: the prediction space of each typical model.
\end{itemize}

\subsubsection{Data Representations}
Table~\ref{tab:sel_task_dr} lists data representations of our two sub-tasks.
For binary models, ``sentence 1'' is the dialogue history, 
while ``sentence 2'' is an entity name or knowledge information, and the target is `True' or `False'.
For the multi-class or list-wise model in knowledge ranking, we put top-$5$ knowledge from a point-wise model in ``sentence 2'' in a batch, 
and minimize the cross-entropy loss between the true distribution and system predictions. 

\subsubsection{Entity Tracking}
Given a complete set of entities $\mathbf{E}$ and a dialogue, entity tracking finds an entity list $E_i$, 
each entity in which is mentioned in the current dialogue. 
The rows 2 to 3 in Table~\ref{tab:sel_task_dr} illustrate the data representation. For each entity $e_j \in \mathbf{E}$, our model 
predicts whether this $e_j$ is mentioned in the current dialogue or not. 
In practice, $e_j$ is mentioned if its score is larger than a threshold $\delta_e$.

Since there are some data sets annotated in a similar way as entity tracking, we extract additional training data from Schema-Guided Dialogue~\cite{sgd}, Topical-Chat~\cite{topical_chat} and Topical-Chat ASR~\cite{topical_chat_asr}.


\subsubsection{Knowledge Ranking}
After we obtain an entity list $E_i$ from entity tracking, we rank them and return a top-$n$ knowledge list. 
We use two types of models: point-wise and list-wise~\cite{listrank}. The point-wise approach scores each candidate independently and ranks them based on their scores, while the list-wise method takes into account all candidates at the same time and the objective is to maximize the probability of the correct one.

\begin{figure}
  \centering
  \includegraphics[width=0.475\textwidth]{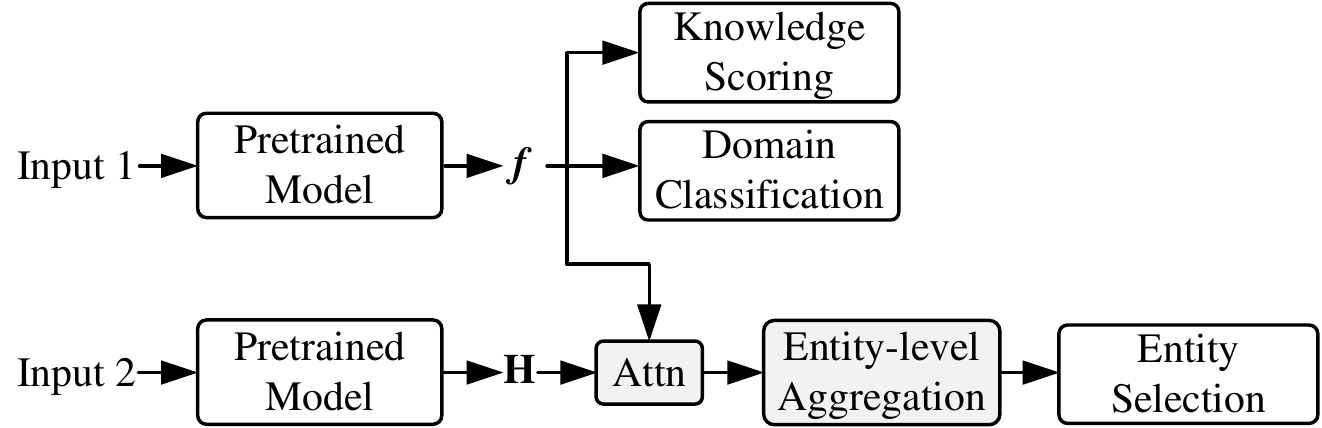}
  \caption{Illustration of the multi-task learning method for knowledge ranking. In addition to the regular knowledge scoring, we design a domain classification task and an entity selection task. Input 1 is the concatenation of Sentence 1 and Sentence 2, and Input 2 is the concatenation of several sampled entity names. $\bm{f}$ denotes the pooling output of hidden states of the encoder for Input 1, and $\mathbf{H}$ denotes the output hidden states of the encoder for Input 2.}
  \label{fig:sel_mtl}
\end{figure}

For the point-wise method, we propose a multi-task learning approach in training to learn multi-level semantic information. 
Typically, we design two auxiliary tasks: domain classification and entity selection in Figure~\ref{fig:sel_mtl}.
Input 1 and Input 2 share a same encoder. 
For domain classification, we use the pooling output of the last hidden states of Input 1 ($\bm{f}$), and predict
the domain of the ground-truth knowledge. 
For entity selection, 
we first sample $n-1$ negative entities from $\mathbf{E}$, then insert the ground-truth entity into a random position and form Input 2 with $n$ entity names. 
Our intuition is that our model should distinguish the correct entity from others according to dialogue utterances.
We run the pre-trained language model over Input 1 and 2 separately, 
and get the last hidden states $\mathbf{H}=\{\bm{h}_i\}_{i=1:T}$ of Input 2, where $T$ is the length of Input 2.
We compute attention scores $\bm{a}$ by using $\bm{f}$ as query and $\mathbf{H}$ as keys. 
Then we have weighted hidden states $\mathbf{G}=\{\bm{g}_i\}_{i=1:T}$ according to Equ.~(\ref{eq:attn}) and Equ.~(\ref{eq:weight}).
\begin{align}
  \bm{a} &= softmax(\frac{1}{\sqrt{d}}\bm{f}\mathbf{W_Q}\mathbf{W_K^T}\mathbf{H}^{T}) \label{eq:attn} \\
  \bm{g}_i &= a_i \bm{h}_i\mathbf{W_V},\ i=1,2,...,T \label{eq:weight}
\end{align}
where $\bm{f}\in \mathbb{R}^{1\times d}, \mathbf{H}\in \mathbb{R}^{T\times d}$, $d$ is the feature dimension. 
$\mathbf{W_Q}, \mathbf{W_K}, \mathbf{W_V}$ are projection matrices. 

To get entity-level features $\mathbf{S}=\{\bm{s}_k\}_{k=1:n}$, we aggregate $\mathbf{G}$ over each entity. 
Let $k_1$ to $k_t$ denote indices of the $k$-th entity in Input 2, where $t$ is the entity length, then we have
\begin{align}
    \bm{s}_k &= \sum_{i=k_1}^{k_t} \bm{g}_i,\ k = 1,2,...,n
\end{align}
Finally, we compute the probability distribution of $n$ entities from $\mathbf{S}$ and minimize the KL divergence between it and the true distribution.

In practice, we sample 3 negative entities for each training instance from entities that appeared in dialogue utterances and entities that belong to the same domain as the ground-truth entity.

We further introduce two binary sparse features to indicate (1) whether the current knowledge is domain-level knowledge and 
(2) whether the entity name of the candidate knowledge is the last entity that appeared in dialogue utterances. 
We simply use string matching to check whether an entity name appear in the dialogue utterances, 
and we integrate sparse features into a Wide \& Deep structure~\cite{chen2016wide}, named \robertawd in our experiments.

\begin{table*}
\begin{center}
\scalebox{0.90}{
\begin{tabular}{ c | c | c | c }
\hline 
\multicolumn{3}{c|}{Context} & Target  \\
\hline
History & Top-$5$ Knowledge & Last Turn & Response \\
\hline
 $\langle user \rangle U_1$  ... $\langle sys \rangle S_{i-1}$  & $\langle kng_5 \rangle \langle ent \rangle e^5_i \langle ans \rangle a^5_i$ ... $\langle kng_1 \rangle \langle ent \rangle e^1_i \langle ans \rangle a^1_i$ & $\langle user \rangle U_{i}$ & $\langle resp \rangle$ $S_i$ \\
\hline
\end{tabular}
}
\end{center}
\abovecaptionskip=3pt 
\caption{The data representations for generation task. }
\label{tab:gen_task}
\end{table*}

For the list-wise method, 
we first run a point-wise model, and get top-$5$ knowledge for each example. Then we apply list-wise models on those 
top-$5$ results (see the last two rows in Table~\ref{tab:sel_task_dr} for an example). 
For the training samples, we use ``{\em k}-fold cross-validated style'' to get system candidate lists. We split the training data into $k$ sub-folds, and train $k$ different models by holding each sub-set as a validation set, then decode each sub-set with the corresponding trained model.

In addition to two sparse features used in point-wise models, 
for list-wise models we add two extra features: whether uni-gram and bi-gram of the current entity name show in the dialogue. 
These features aim to help the model better deal with the challenge that the entity name is scattered in the dialogue utterances. 
For example, for the entity ``Hilton San Francisco Union Square'', the words ``San Francisco'' and ``Union Square'' may be mentioned at the beginning of the conversation,
while ``Hilton'' may appear in the end. 

Our ensemble re-ranks all knowledge by using the sum of probabilities of each knowledge from all single systems. 

\subsection{Generation}\label{sec:gen}
Following Mi et al.~\shortcite{mi2021towards}, 
we make use of pre-trained language models, such as UniLM~\cite{unilm}, GPT2~\cite{gpt2} and DialoGPT~\cite{dialogpt} for our generation task, 
and adopt their online training and ensemble methods. For more details, please refer to Mi et al.~\shortcite{mi2021towards}. 
Additionally, we perform extra data processing on DSTC-10 data in order to capture the characteristics of spoken conversations.

\subsubsection{Data Preprocessing}
Besides the usage of data augmentation in Section~\ref{sec:da}, we also modify the response style in our training data in order to match the style of DSTC-10 validation set.
Typically, the response in DSTC-9 always ends with different types of interrogative sentences, e.g. ``Would you like to book a room?'', which is not shown in DSTC-10 validation.
Thus, 
we first collect some high-frequency interrogative sentences, then delete them from our training data.

\subsubsection{Models and Data Representations}
Table~\ref{tab:gen_task} shows the data representations for generation task, where top-$5$ knowledge are given in a descending order from $5$th to $1$st.
With those representations, we fine-tune UniLM~\cite{unilm}, GPT2~\cite{gpt2} and DialoGPT~\cite{dialogpt}
on the training set, and pick the best models based on the scores on the validation set.

\section{Experiments}
\subsection{Data and Common Settings}
\label{sec:data}


We use the DSTC-9 dataset, including training set, validation set, and test set as the original training data, 
then we perform data augmentation on the original training data. We evaluate our systems on the DSTC-10 validation set.



For all data representations, if a context length is larger than the maximum block size, we always trunk the left-most part of the context.
``{\em k}-fold cross-validated style'' means that we first split a data into $k$ sub-folds, and train $k$ different models by holding each sub-set as a validation set, then we decode each sub-set with the corresponding trained model.

\subsection{Detection}
We fine-tune pre-trained language models on our training data 
for $10$ epochs with a learning rate of $1\mathrm{e}{-5}$. The batch size is $16$ for \roberta related models and $32$ for ELECTRA related models.
The optimizer is AdamW.

\begin{table}
    \centering
    \scalebox{0.95}{
    \begin{tabular}{l|c|c|c|c}
        \hline
         System     & DA  & Precision & Recall & F1 \\
        \hline
        \roberta-256 & No & 1.0 & 0.6538 & 0.7907 \\
        \roberta-256 & Yes & 0.9900 & 0.9230 & 0.9553 \\
        \roberta-512 & Yes & 0.9800 & 0.9420 & 0.9606 \\
        ELECTRA-256 & Yes & 1.0    & 0.9615 & 0.9804 \\
        ELECTRA-512 & Yes & 0.9717 & 0.9904 & 0.9810 \\
        \hline \hline
        Ensemble  & Yes & 0.9900 & 0.9810 & 0.9855 \\
        \hline
    \end{tabular}
    }
    \abovecaptionskip=5pt 
    \caption{Detection results on the validation set. ``DA'' means data augmentation. -256 and -512 mean the maximum context length. The \roberta-256 model without data augmentation is trained on the DSTC-9 training set.}
    \label{tab:det_ens_val}
\end{table}


Table~\ref{tab:det_ens_val} lists results of different models on the validation set. 
It is clear that training on written conversations fails to perform well on spoken conversations. 
After we run data augmentation, we see significant improvements in terms of F1 score, and longer context length yields slightly better F1. 
Those results also suggest that ELECTRA models always perform better than \roberta in terms of F1 score. 
For the error-fixing ensemble, we select the ELECTRA-512 as our base model. The threshold of $\delta_d$ is $0.3$, and our ensemble achieves the best F1 score at $0.9855$.

\begin{table}
    \centering
    \scalebox{0.95}{
    \begin{tabular}{l|c|c|c}
        \hline
         System       & Precision & Recall & F1 \\
        \hline
        Baseline: DSTC-9   & 0.9017    & 0.7116 & 0.7954 \\
        Baseline: Knover   & 0.8967    & 0.6735 & 0.7692 \\
        \hline \hline
        Ensemble & 0.8814 & 0.9575 & 0.9179 \\
        \hline
    \end{tabular}
    }
    \abovecaptionskip=5pt 
    \caption{Detection results on the test set.}
    \label{tab:det_ens_test}
\end{table}

Table~\ref{tab:det_ens_test} shows the results on the test set, with data augmentation and ensemble algorithms, 
our model achieves 0.9179 in terms of F1 score.

\subsection{Selection}
For the selection task, we compare different entity tracking methods, data augmentation, negative sampling strategies, and model structures.
The $\delta_e$ for entity tracking threshold is $0.5$.
We train all models on the training data for $2$ epochs with a learning rate of $1\mathrm{e}{-5}$, a batch size of $16$, and the optimizer of AdamW. 
For a fair comparison, we use the ground truth of detection for experiments on the validation set. 
As our training set includes DSTC-9 test set, and a portion of DSTC-9 test set also shows in DSTC-10 validation set but with spoken language, 
experimental results on the validation set are relatively high. 
Please also note that we add DSTC-10 validation set into our training for the final DSTC-10 test set.

\subsubsection{Entity Tracking}
\begin{table}
\centering
    \scalebox{0.95}{
    \begin{tabular}{l|c|ccc}
    \hline
    \multirow{2}{*}{System}      & Entity Tra. & \multicolumn{3}{c}{Knowledge Ranking}  \\ 
    \cline{2-5}
                & Recall          & MRR@5  & R@1    & R@5                  \\ 
    \hline
    Exact Match & 0.8077          & 0.6566 & 0.6154 & 0.7115               \\
    Fuzzy Match & 0.9808          & 0.7662 & 0.7115 & 0.8558               \\
    \roberta    & 0.9904          & 0.8075 & 0.7308 & 0.9231               \\
    \hline
    \end{tabular}
    }
    \caption{Comparison of different entity tracking methods on the validation set. The exact/fuzzy match methods check whether an entity name shows in dialogue utterances by exact/fuzzy matching.}
    \label{tab:entity_tracking}
\end{table}

Table~\ref{tab:entity_tracking} shows the entity tracking and the corresponding knowledge ranking results on the validation set. 
The exact and fuzzy match approaches directly check whether an entity name exists in dialogue utterances by exact and fuzzy matching, respectively.
Please note that the \roberta model is trained directly on training data with data augmentation
(including more data from Schema-Guided Dialogue~\cite{sgd}, Topical-Chat~\cite{topical_chat} and Topical-Chat ASR~\cite{topical_chat_asr}, and data augmentation methods in Section~\ref{sec:da}).
The knowledge ranking model is a point-wise \robertawd without multi-task learning and trained on the original training set without data augmentation. 
Those results suggest that it is unsatisfactory to use exact match method due to the omission of a part of entity name and speech recognition errors, which can be alleviated by fuzzy matching. 
However, fuzzy matching may introduce considerable candidate entities and bring difficulties for the subsequent knowledge ranking task. 
Instead, using a \roberta model not only achieves the highest recall rate on the entity tracking task, but also improves the performance on the knowledge ranking task.

\subsubsection{Data Augmentation}
\begin{table}
\centering
\scalebox{0.95}{
    \begin{tabular}{l|c|c|c}
    \hline
    Data              & MRR@5  & R@1    & R@5     \\ 
    \hline
    \robertawd        & 0.8075 & 0.7308 & 0.9231  \\
    + AEI             & 0.8365 & 0.7596 & 0.9519  \\
    + AEI + TST       & 0.8564 & 0.7981 & 0.9327  \\
    + AEI + TST + ENA & 0.8838 & 0.8173 & 0.9712  \\
    \hline
    \end{tabular}
    }
    \abovecaptionskip=5pt
    \caption{The effect of data augmentation methods on the validation set. ``AEI'', ``TST'', and ``ENA'' denote artificial error injection, text-speech-text transformation, and entity name augmentation, respectively.}
    \label{tab:sel_data_aug}
\end{table}

\begin{table}
\centering
\scalebox{0.95}{
    \begin{tabular}{l|c|c|c}
    \hline
    Model                & MRR@5  & R@1    & R@5     \\ 
    \hline
    \robertawd*           & 0.8822 & 0.8365 & 0.9615  \\
    \robertawdmtl       & 0.8966 & 0.8462 & 0.9712  \\
    \robertawd-listwise  & 0.9006 & 0.8558 & 0.9712  \\
    \robertawdd-listwise & 0.9290 & 0.9038 & 0.9712  \\
    \hline
    \end{tabular}
    }
    \abovecaptionskip=5pt
    \caption{Results of different models on the validation set.
    \robertawd* is the point-wise model with data augmentation and better negative sampling.
    \robertawdmtl\ adds multi-task learning. \robertawdd-listwise uses two additional sparse features to indicate whether the uni-gram and bi-gram of the current entity name appeared in the dialogue history.}
    \label{tab:sel_model}
\end{table}

\begin{figure}
  \centering
  \includegraphics[width=0.45\textwidth]{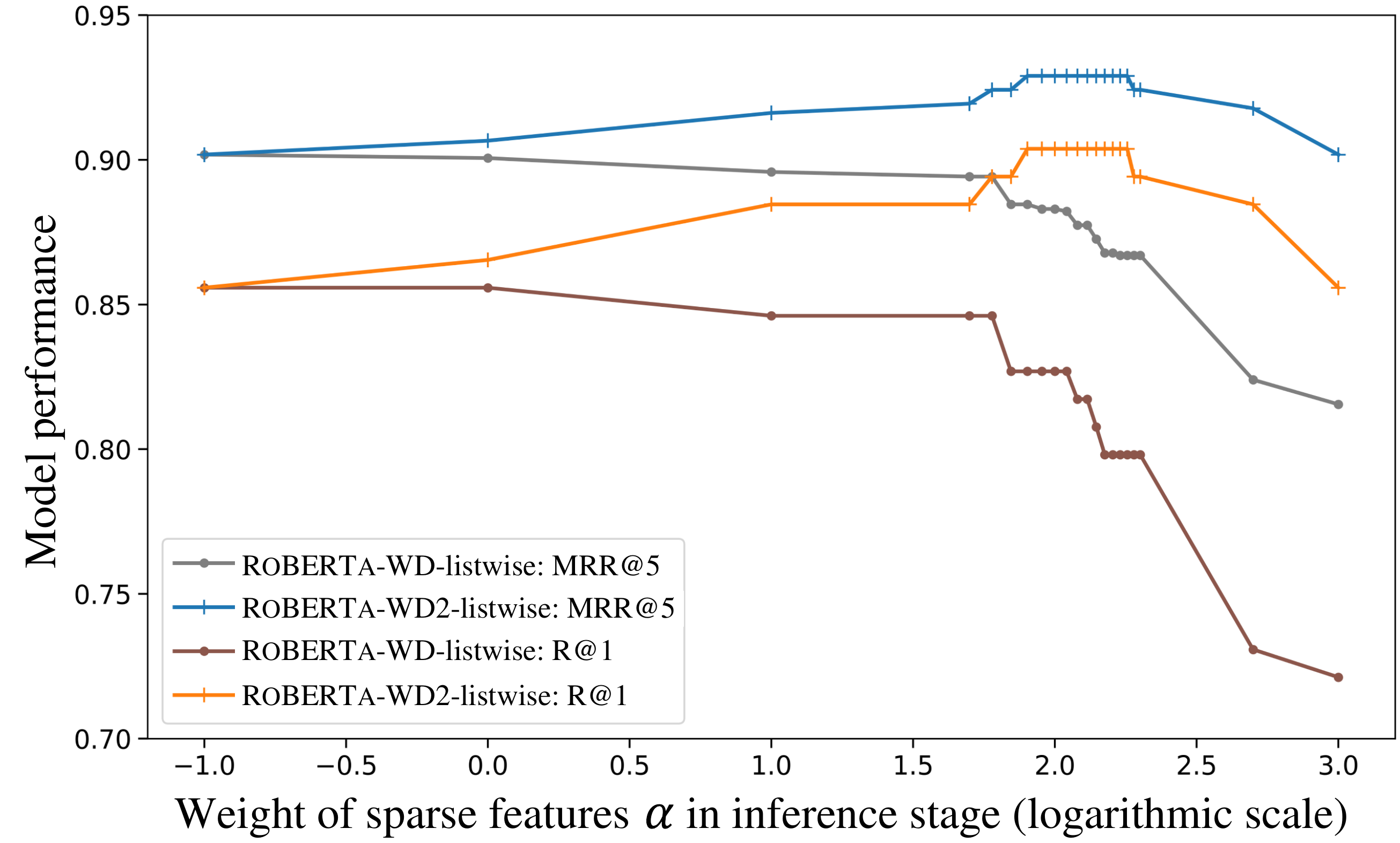}
  \abovecaptionskip=5pt
  \caption{The influence of the weight of sparse features in inference stage on the performance for list-wise models. We only show R@1 and MRR@5 because R@5 remains unchanged for all list-wise models.}
  \label{fig:sel_wd}
\end{figure}

\begin{table}
\centering
\scalebox{0.95}{
    \begin{tabular}{l|c|c|c}
    \hline
    Model                & MRR@5  & R@1    & R@5     \\ 
    \hline
    Baseline: DSTC-9     & 0.5230 & 0.4583 & 0.6252 \\
    Baseline: Knover     & 0.5574 & 0.4950 & 0.6472 \\
    \hline \hline
    \robertawdd-listwise & 0.7541 & 0.7004 & 0.8225 \\
    \hline
    Ensemble             & 0.7891 & 0.7481 & 0.8407 \\
    \hline
    \end{tabular}
    }
    \abovecaptionskip=5pt
    \caption{Selection results on the test set. \robertawdd-listwise is our best single model trained on the training set, and our ensemble model further contains models trained on both training set and DSTC-10 validation set.}
    \label{tab:sel_test}
\end{table}

Table~\ref{tab:sel_data_aug} shows the knowledge ranking results of a basic point-wise \robertawd model with different levels of data augmentation. 
``AEI'', ``TST'', and ``ENA'' denote artificial error injection, text-speech-text transformation, and entity name augmentation, respectively.
By adding more and more data augmentation methods, we see consistent and significant improvements in terms of R@1.
Although text-speech-text transformation degenerates the R@5 score, it boosts R@1 and MRR@5 significantly. 
Together with all three data augmentation methods, we improve MRR@5 from 0.8075 to 0.8838, R@1 from 0.7308 to 0.8173, and R@5 from 0.9231 to 0.9712.

\subsubsection{Negative Sampling} 
In the original knowledge ranking system of Mi et al.~\shortcite{mi2021towards}, 
the negative training instances for point-wise knowledge ranking models are equally sampled from (1) the complete knowledge base and (2) knowledge about entities that appeared in dialogue utterances. 
However, we found that the knowledge of non-ground-truth entities in the conversation is more likely to be recognized as a false positive candidate than knowledge of a ground-truth entity. 
Thus, we extend their negative sample candidates
by adding a knowledge from a non-ground-truth entity that also shows in a conversation.
This negative sampling strategy significantly improves R@1 from 0.8173 further to 0.8365.

\begin{table*}
\begin{center}
\scalebox{0.95}{
\begin{tabular}{  c | c | c | c | c | c | c | c |c |c}
\hline 
Type                    & System & \bleu-1 & \bleu-2 & \bleu-3 & \bleu-4 & \meteor & \rouge-1 & \rouge-2 & \rouge-L \\
\hline
\multirow{4}{*}{Single} & UniLM Uncased  & 0.1509 & 0.0703 & 0.0101 & 0.0125 & 0.1532 & 0.2007 & 0.0470 & 0.1188 \\
                        & UniLM Large    & 0.1909 & {0.1103} & {0.0501} & {0.0204} & {0.2140} & 0.2501 & {0.0970} & {0.1689} \\
                        & GPT2           & 0.1628 & {0.0966} & {0.0544} & {0.0271} & {0.1839} & 0.2092 & {0.0859} & {0.1406} \\
                        & DialoGPT       & 0.1544 & 0.0813 & 0.0452 & 0.0169 & 0.1655 & 0.1840 & 0.0721 & 0.1396 \\
\hline
\multirow{3}{*}{Ensemble} & 10 features       & 0.2072 & {0.1435} & {0.0980} & {0.0560} & {0.1834} & 0.2501 & {0.1334} & {0.2172} \\
& + DA  & {0.3479} & 0.2512 & 0.1658 & 0.1260 & 0.3885 & {0.3735} & 0.2061 & 0.3267 \\
& + DA  \& DP & {0.3613} & 0.2932 & 0.2405 & 0.1871 & 0.4295 & {0.4170} & 0.2816 & 0.3670 \\
\hline
\end{tabular}
}
\end{center}
\abovecaptionskip=3pt
\caption{Validation results of top four single systems (ranked by \bleu-4), and consensus decoding with 10 basic features and data augmentation. 
$10$ basic features include $9$ similarity features, and $1$ reciprocal rank feature \cite{mi2021towards}.
``DA'' means data augmentation methods shown in Section~\ref{sec:da}, and ``DP'' means data preprocessing of removing high frequent interrogative sentences from responses.}
\label{tab:validation-results}
\end{table*}

\subsubsection{Comparison of different models} 

Table~\ref{tab:sel_model} lists the results of different models on the validation set. 
For point-wise models, with the help of multi-task learning method, our \robertawdmtl improves all metrics by about 1 percent.
For list-wise models, \robertawd-listwise system uses two binary sparse features to indicate the domain of the current knowledge and whether the current entity is the last entity in the conversation. 
The list-wise ranking helps about another 1 percent in terms of R@1. 
The \robertawdd-listwise system further adds two additional sparse features to identify whether the uni-gram and bi-gram of the current entity name appeared in the dialogue history, 
and improves the R@1 by almost 5\%.

In order to better utilize sparse features,  
we scale sparse feature indicator by a weight $\alpha$, which means that the input of a fired sparse feature is $\alpha$ instead of 1.
Figure~\ref{fig:sel_wd} shows the performance curves with different $\alpha$ on the validation set.
It is interesting that as $\alpha$ goes up, \robertawd-listwise gets worse, while \robertawdd-listwise performs better before reaching a threshold.
It suggests that increasing uni-gram and bi-gram feature weights to a range from 80 to 180 leads to the best performance on the validation set.
Thus, we fix $\alpha$ to be 100 in inference stage, 
our best single model achieves 0.9038 in terms of R@1 (the last row in Table~\ref{tab:sel_model}).

\subsubsection{Final Results on Test}
Table~\ref{tab:sel_test} presents results of our best single model (\robertawdd-listwise) and ensemble model on the final test set. \robertawdd-listwise is trained on the training set, and our ensemble model further contains models trained on both training set and DSTC-10 validation set.
The proposed techniques, including data augmentation, multi-task learning, artificial sparse features, and ensemble algorithm bring significant performance gains over baseline systems.

\subsection{Generation}
We fine-tune pre-trained models with batch size $32$ for $6$ epochs over the training data. 
We use three UniLM models\footnote{https://github.com/microsoft/unilm/tree/master/s2s-ft}.
For the UniLM Large models, the maximum history size of each context is $512$, and the maximum target size is $96$. 
The optimizer is AdamW with a learning rate of $1\mathrm{e}{-5}$, weight decay of $0.01$.
UniLM Uncased models use $640$ as the maximum history size. 
The $p_s$ of UniLM Large is $0.15$.
GPT2 and DialoGPT models use a learning rate of $5\mathrm{e}{-5}$, and the maximum block size is $512$. 
We use base models for GPT2 and DialoGPT.


\begin{table}
\begin{center}
\scalebox{0.95}{
\begin{tabular}{c|c|c|c}
\hline 
Type & Sys & \bleu-1 & \bleu-4  \\
\hline 
\multicolumn{2}{c|}{Baseline: DSTC-9} & 0.1153  &	0.0075 \\
\multicolumn{2}{c|}{Baseline: Knover} & 0.1248  &	0.0153 \\
\hline
\hline
\multirow{1}{*}{Single}  & UniLM Large     & {0.1802}  & 0.0460 \\
\hline
\multirow{3}{*}{Ensemble}  & 10 features     & {0.3150}  & 0.0732  \\
                      & + DA                 & {0.3228}  & 0.1019  \\
                      & + DA \& DP            & 0.3401 & {0.1154}  \\
\hline
\end{tabular}
}
\end{center}
\abovecaptionskip=3pt
\caption{Test results of our models. We only show \bleu-1 and \bleu-4 due to space limitation.}
\label{tab:test-results}
\end{table}

\subsubsection{Discrepancy in Pipeline Framework}
Following Mi et al.~\shortcite{mi2021towards}, we decode the training set in ``{\em 10}-fold cross-validated style'' to alleviate the discrepancy in the pipeline framework.
The \rone score of the decoded training set is about 0.9.
We then train our model on the decoded training set, and observe 1 percent improvement in terms of \bleu-4.

\subsubsection{Results}
Table~\ref{tab:validation-results} shows top four single systems and ensemble results on the validation set.
UniLM Large uses top-$5$ knowledge in context, and performs best on 6 of 8 metrics. 
UniLM Uncased achieves comparable results to UniLM Large.
Please note that, all the single systems shown in Table~\ref{tab:validation-results} only use DSTC 9 training set without any data augmentation.
Our consensus decoding with $10$ basic features improves \bleu-4 scores by $3.6$ percent on the validation set,
After we perform data augmentation in Section~\ref{sec:da}, the \bleu-4 scores are improved by $7$ percent, and the data preprocessing by removing high frequent interrogative sentences further improves \bleu-4 scores by $6.1$ percent.

Table~\ref{tab:test-results} shows the results on the final test set.
Our single model UniLM Large is significantly better than official baseline systems, 
and consensus decoding improves \bleu-4 by $2.7$ points.
Adding DA and DP further improve \bleu-4 scores by $2.8$ and $1.4$ percent, respectively. 
Those improvements are smaller than the improvements on our validation set due to the following two facts:
1) the test set is about 10 times larger than validation set; 
2) the validation set includes partial DSTC 9 test set, which is used in our training.

\subsection{Human Evaluation Results}
Table~\ref{tab:human_results} shows the official human evaluation results on the test set, 
`appropriateness' measures how well a system output is naturally connected to a given conversation, 
while `accuracy' measures the accuracy of a system output against the reference knowledge.
Our ensemble system ranks in the second place.

\begin{table}
    \centering
    \scalebox{0.95}{
    \begin{tabular}{c|c|c|c}
        \hline
         System       & Accuracy & Appropriateness & Average \\
        \hline
        Ground-truth      & 3.5769 & 3.4814 & 3.5292 \\
        \hline \hline
        Best & 3.4947 & 3.3523 & 3.4235 \\
        Ours & 3.3356 & 3.3021 & 3.3189 \\
        \hline
    \end{tabular}
    }
    \abovecaptionskip=3pt
    \caption{Human evaluation results on the test set. }
    \label{tab:human_results}
\end{table}

\section{Conclusion}
In this paper, we have presented our system pipeline for the sub-track 2 of ``Knowledge-grounded Task-oriented Dialogue Modeling on Spoken Conversations'' in DSTC-10.
To bridge the gap between speaking and writing, we adopt extensive data augmentation methods. 
We also use multi-task learning, and different ensemble algorithms for different sub-tasks to train robust models with high generalization ability.
Overall, experimental results show that data augmentation significantly boosts model performance on all three sub-tasks, 
the multi-task learning enhances the model's ability to learn multi-level semantic information on the knowledge ranking task,
and all ensemble algorithms improve the final objective metrics.
Our approach has ranked third on objective metrics and second on human evaluation. In future work, we will further improve the generalization ability of our models by generating more spoken-like training data and increasing model diversity.


\bibliography{main}

\begin{thebibliography}{19}
\providecommand{\natexlab}[1]{#1}

\bibitem[{Amodei et~al.(2016)Amodei, Ananthanarayanan, Anubhai
  et~al.}]{amodei2016deep}
Amodei, D.; Ananthanarayanan, S.; Anubhai, R.; et~al. 2016.
\newblock {Deep Speech 2: End-to-End Speech Recognition in English and
  Mandarin}.
\newblock In \emph{ICML}.

\bibitem[{Cao et~al.(2007)Cao, Qin, Liu, Tsai, and Li}]{listrank}
Cao, Z.; Qin, T.; Liu, T.-Y.; Tsai, M.-F.; and Li, H. 2007.
\newblock Learning to rank: from pairwise approach to listwise approach.
\newblock In \emph{Proceedings of the 24th international conference on Machine
  learning}, ICML '07, 129--136. New York, NY, USA: ACM.
\newblock ISBN 978-1-59593-793-3.

\bibitem[{Cheng et~al.(2016)Cheng, Koc, Harmsen et~al.}]{chen2016wide}
Cheng, H.-T.; Koc, L.; Harmsen, J.; et~al. 2016.
\newblock {Wide \& Deep Learning for Recommender Systems}.
\newblock In \emph{DLRS 2016}, 7–10.

\bibitem[{Clark et~al.(2020)Clark, Luong, Le, and Manning}]{clark2020electra}
Clark, K.; Luong, M.-T.; Le, Q.~V.; and Manning, C.~D. 2020.
\newblock {ELECTRA}: Pre-training Text Encoders as Discriminators Rather Than
  Generators.
\newblock In \emph{ICLR}.

\bibitem[{Dong et~al.(2019)Dong, Yang, Wang, Wei, Liu, Wang, Gao, Zhou, and
  Hon}]{unilm}
Dong, L.; Yang, N.; Wang, W.; Wei, F.; Liu, X.; Wang, Y.; Gao, J.; Zhou, M.;
  and Hon, H. 2019.
\newblock Unified Language Model Pre-training for Natural Language
  Understanding and Generation.
\newblock \emph{CoRR}, abs/1905.03197.

\bibitem[{Gopalakrishnan et~al.(2019)Gopalakrishnan, Hedayatnia, Chen,
  Gottardi, Kwatra, Venkatesh, Gabriel, and Hakkani-Tür}]{topical_chat}
Gopalakrishnan, K.; Hedayatnia, B.; Chen, Q.; Gottardi, A.; Kwatra, S.;
  Venkatesh, A.; Gabriel, R.; and Hakkani-Tür, D. 2019.
\newblock {Topical-Chat: Towards Knowledge-Grounded Open-Domain Conversations}.
\newblock In \emph{Proc. Interspeech 2019}, 1891--1895.

\bibitem[{Gopalakrishnan et~al.(2020)Gopalakrishnan, Hedayatnia, Wang, Liu, and
  Hakkani-Tür}]{topical_chat_asr}
Gopalakrishnan, K.; Hedayatnia, B.; Wang, L.; Liu, Y.; and Hakkani-Tür, D.
  2020.
\newblock {Are Neural Open-Domain Dialog Systems Robust to Speech Recognition
  Errors in the Dialog History? An Empirical Study}.
\newblock In \emph{INTERSPEECH}.

\bibitem[{He et~al.(2021)He, Lu, Bao, Wang, Wu, Niu, and Wang}]{he2021learning}
He, H.; Lu, H.; Bao, S.; Wang, F.; Wu, H.; Niu, Z.; and Wang, H. 2021.
\newblock {Learning to Select External Knowledge with Multi-scale Negative
  Sampling}.
\newblock \emph{AAAI DSTC9 Workshop}.

\bibitem[{Kim et~al.(2020)Kim, Eric, Gopalakrishnan, Hedayatnia, Liu, and
  Hakkani-Tur}]{kim2020domain}
Kim, S.; Eric, M.; Gopalakrishnan, K.; Hedayatnia, B.; Liu, Y.; and
  Hakkani-Tur, D. 2020.
\newblock Beyond Domain APIs: Task-oriented Conversational Modeling with
  Unstructured Knowledge Access.
\newblock arXiv:2006.03533.

\bibitem[{Kim et~al.(2021)Kim, Eric, Hedayatnia, Gopalakrishnan, Liu, Huang,
  and Hakkani-Tur}]{kim2021beyond}
Kim, S.; Eric, M.; Hedayatnia, B.; Gopalakrishnan, K.; Liu, Y.; Huang, C.-W.;
  and Hakkani-Tur, D. 2021.
\newblock {Beyond Domain APIs: Task-oriented Conversational Modeling with
  Unstructured Knowledge Access Track in DSTC9}.
\newblock \emph{arXiv preprint arXiv:2101.09276}.

\bibitem[{Liu et~al.(2019)Liu, Ott, Goyal, Du, Joshi, Chen, Levy, Lewis,
  Zettlemoyer, and Stoyanov}]{roberta}
Liu, Y.; Ott, M.; Goyal, N.; Du, J.; Joshi, M.; Chen, D.; Levy, O.; Lewis, M.;
  Zettlemoyer, L.; and Stoyanov, V. 2019.
\newblock RoBERTa: {A} Robustly Optimized {BERT} Pretraining Approach.
\newblock \emph{CoRR}, abs/1907.11692.

\bibitem[{Meechan-Maddon(2019)}]{ailsa2019effect}
Meechan-Maddon, A. 2019.
\newblock {The effect of noise in the training of convolutional neural networks
  for text summarisation}.

\bibitem[{Mi et~al.(2021)Mi, Ren, Dai, He, Sun, Li, Zheng, and
  Xu}]{mi2021towards}
Mi, H.; Ren, Q.; Dai, Y.; He, Y.; Sun, J.; Li, Y.; Zheng, J.; and Xu, P. 2021.
\newblock {Towards Generalized Models for Beyond Domain API Task-oriented
  Dialogue}.
\newblock \emph{AAAI DSTC9 Workshop}.

\bibitem[{Pauls, DeNero, and Klein(2009)}]{pauls2009consensus}
Pauls, A.; DeNero, J.; and Klein, D. 2009.
\newblock {Consensus Training for Consensus Decoding in Machine Translation}.
\newblock In \emph{EMNLP}, 1418--1427.

\bibitem[{Radford et~al.(2019)Radford, Wu, Child, Luan, Amodei, and
  Sutskever}]{gpt2}
Radford, A.; Wu, J.; Child, R.; Luan, D.; Amodei, D.; and Sutskever, I. 2019.
\newblock Language Models are Unsupervised Multitask Learners.

\bibitem[{Rastogi et~al.(2020)Rastogi, Zang, Sunkara, Gupta, and Khaitan}]{sgd}
Rastogi, A.; Zang, X.; Sunkara, S.; Gupta, R.; and Khaitan, P. 2020.
\newblock {Towards Scalable Multi-domain Conversational Agents: The
  Schema-Guided Dialogue Dataset}.
\newblock arXiv:1909.05855.

\bibitem[{Shen et~al.(2018)Shen, Pang, Weiss, Schuster, Jaitly, Yang, Chen,
  Zhang, Wang, Skerry-Ryan, Saurous, Agiomyrgiannakis, and
  Yonghui}]{shen2018natural}
Shen, J.; Pang, R.; Weiss, R.~J.; Schuster, M.; Jaitly, N.; Yang, Z.; Chen, Z.;
  Zhang, Y.; Wang, Y.; Skerry-Ryan, R.; Saurous, R.~A.; Agiomyrgiannakis, Y.;
  and Yonghui, W. 2018.
\newblock {Natural TTS Synthesis By Conditioning Wavenet On Mel Spectrogram
  Predictions}.
\newblock In \emph{ICASSP}.

\bibitem[{Tang et~al.(2021)Tang, Shang, Lv, Fu, Zhang, Huang, and
  Zhang}]{tang2021radge}
Tang, L.; Shang, Q.; Lv, K.; Fu, Z.; Zhang, S.; Huang, C.; and Zhang, Z. 2021.
\newblock {RADGE: Relevance Learning and Generation Evaluating Method for
  Task-oriented Conversational System}.
\newblock \emph{AAAI DSTC9 Workshop}.

\bibitem[{Zhang et~al.(2020)Zhang, Sun, Galley, Chen, Brockett, Gao, Gao, Liu,
  and Dolan}]{dialogpt}
Zhang, Y.; Sun, S.; Galley, M.; Chen, Y.-C.; Brockett, C.; Gao, X.; Gao, J.;
  Liu, J.; and Dolan, B. 2020.
\newblock DialoGPT: Large-Scale Generative Pre-training for Conversational
  Response Generation.
\newblock In \emph{ACL, system demonstration}.

\end{thebibliography}


\end{document}